\pdfoutput=1
\documentclass[lettersize,journal]{IEEEtran}
\usepackage{amsmath,amsfonts}
\usepackage{algorithmic}
\usepackage{algorithm}
\usepackage{array}
\usepackage[caption=false,font=normalsize,labelfont=sf,textfont=sf]{subfig}

\def\BibTeX{{\rm B\kern-.05em{\sc i\kern-.025em b}\kern-.08em
		T\kern-.1667em\lower.7ex\hbox{E}\kern-.125emX}}
	
\usepackage{textcomp}
\usepackage{stfloats}
\usepackage{url}
\usepackage{verbatim}
\usepackage{graphicx}
\usepackage{cite}

\usepackage{hyperref}
\usepackage{lineno}
\usepackage{caption}

\usepackage{threeparttable}

\usepackage{pifont}
\usepackage{url}

\usepackage{tabu}                     
\usepackage{multirow}                 
\usepackage{multicol}                 
\usepackage{multirow}                
\usepackage{float}                    
\usepackage{makecell}                 
\usepackage{booktabs}                 

\usepackage[none]{hyphenat}
\usepackage{amssymb}
\usepackage{diagbox}
\usepackage{bbding}

\usepackage{algorithm}
\usepackage{algorithmic}

\usepackage{color}
\usepackage{xcolor}
\usepackage{colortbl, booktabs}
\usepackage{graphicx}

\begin{document}

\title{AMIEOD: Adaptive Multi-Experts Image Enhancement for Object Detection in Low-Illumination Scenes }

        \author{Xiaochen Huang, 
        	  Honggang Chen,~\IEEEmembership{Member, ~IEEE,}
        		Weicheng Zhang, 
        		Xiaobo Dai,
        		Yongyi Li,
        		\\Linbo Qing,~\IEEEmembership{Member, ~IEEE,}
        		Xiaohai He,~\IEEEmembership{Member, ~IEEE}

\thanks{
			This work was supported by the Open Fund of Key Laboratory of the Ministry of Education on Artificial Intelligence in Equipment [grant number 2024-AAIE-KF04-03]; 
			by the Science and Technology Projects of Xizang Autonomous Region [grant number XZ202501ZY0064]; 
			by the Police Integration Computing Key Laboratory of Sichuan Province [grant number No. JWRH202502002]; 
			and by the TCL Science and Technology Innovation Fund.(Corresponding author: Honggang Chen).
			}
\thanks{
		Xiaochen Huang,  Weicheng Zhang, Xiaobo Dai, Yongyi Li, Linbo Qing and Xiaohai He are with the College of Electronics and Information Engineering, Sichuan University, Chengdu 610065, China; (e-mail: xiaochen\_huang@stu.scu.edu.cn; wczhang21@iflytek.com; 2023222055133 @stu.scu.edu.cn; Yongyi\_Li@stu.scu.edu.cn; qing\_lb@scu.edu.cn; hxh@scu. edu.cn).
		}
\thanks{
		Honggang Chen is with the College of Electronics and Information Engineering, Sichuan University, Chengdu 610065, China, also with the Police Integration Computing Key Laboratory of Sichuan Province, China (e-mail: honggang\_chen@scu.edu.cn).
		}
}

\markboth{Manuscript Submitted to IEEE Transactions on Multimedia, Oct., 2025}%
{Huang \MakeLowercase{\textit{\textit{et al}.}}: AMIEOD: Adaptive Multi-Experts Image Enhancement for Object Detection in Low-Illumination Scenes}


\maketitle

\begin{abstract}
	\sloppy{}
	In multimedia application scenarios, images captured under low-illumination conditions often lead to lower accuracy in visual perception tasks compared to those taken in well-lit environments. 
	To tackle this challenge, we propose AMIEOD, an image enhancement-enabled object detection framework for low-illumination scenes, where the two tasks are jointly optimized in a detection performance-oriented manner.
	Specifically, to fully exploit the information in poorly lit images, a Multi-Experts Image Enhancement Module (MEIEM) is proposed, which leverages diverse enhancement strategies. 
	On this basis, aiming to better align the MEIEM with the detection task, we propose a Detection-Guided Regression Loss (DGRL) that utilizes the detection result to decide the regression target. 
	Moreover, to dynamically select the most suitable enhancement strategy from MEIEM during inference, we construct an Expert Selection Module (ESM) guided by the proposed Detection-Guided Cross-Entropy (DGCE) loss, which formulates the optimization of ESM as a classification task. 
	The improved method is well-matched with current detection algorithms to improve their performance in dim scenes.
	Extensive experiments on multiple datasets demonstrate that the proposed method significantly improves object detection accuracy in low-illumination conditions. Our code has been released at https://github.com/scujayfantasy/AMIEOD
\end{abstract}

\begin{IEEEkeywords}
Cross-resolution knowledge distillation, remote sensing images, object detection, feature enhancement.
\end{IEEEkeywords}


\section{Introduction \label{introduction}}
\sloppy{}
\IEEEPARstart{O}{bject} detection is one of the fundamental tasks in computer vision \cite{chen2024review,dai2023adaptive,11154015}, having achieved prominent achievements and been widely applied in various fields such as autonomous driving, security surveillance, etc\cite{wang2024degradation, li2023adaptive}. 
However, existing learning-based object detection models have certain limitations when directly applied in practical scenarios\cite{HU2026132726}, because they are primarily trained on standard image datasets \cite{lin2014microsoft, hoiem2009pascal}, and these images are usually collected under relatively ideal conditions.
When applied to low-illumination images such as those captured at night, backlighting, or short exposure conditions, their detection performance drops significantly, which in turn affects the decision-making capability and accuracy of the decision systems. 
\par 
In practice, effective intelligent monitoring systems for poorly lit images are both essential and critical.
Therefore, some researches \cite{ma2022toward,sun2022rethinking,yin2023pe,qin2022denet,liu2022image,kalwar2023gdip,ogino2025erup, feng2024toward, feng2025lightstar, ji2025fcma,cui2024trash, cui2021multitask,kennerley20232pcnet,zhang2024isp,du2024boosting,hong2024you} concentrate on the improving the robustness and generalization capability of detection algorithms under low-illumination conditions, which holds considerable practical relevance.
\par 
    \begin{figure}[t]
    	\centering
    	\includegraphics[scale=0.18]{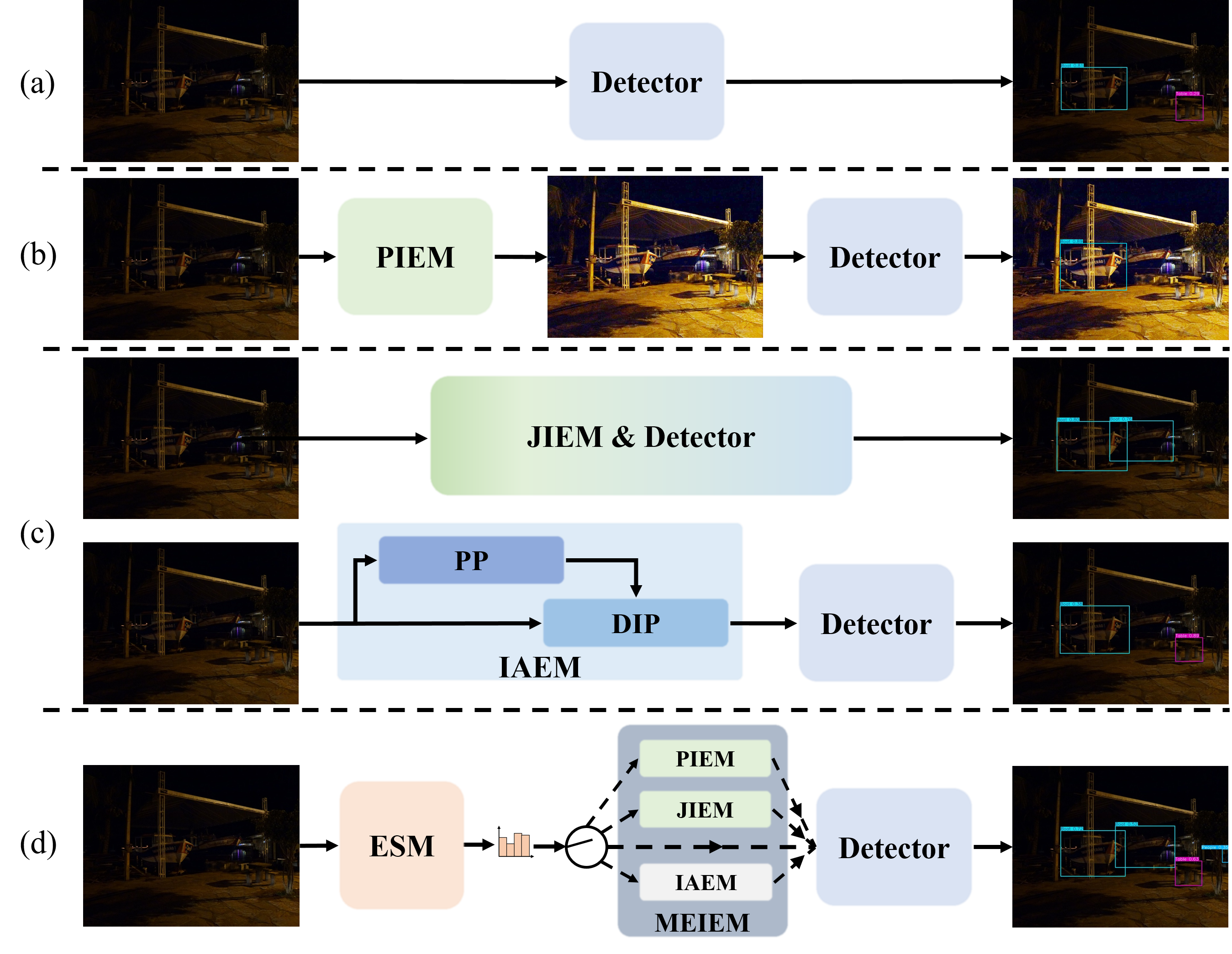}
    	\caption{General structure of different low-illumination image detection algorithms. (a) Directly detecting; (b) LLIE; (c) JED; (d) Our AMIEOD. ``PP'' indicates the Parameter Prediction module. ``DIP'' is the Differentiable Image Processing module.}
    	\label{f1}
    \end{figure}
    Most of object detection methods for low-illumination images can be classified into three categories, namely, \textbf{LLIE}-based methods (Low-Light Image Enhancement)\cite{lv2018mbllen,zhang2019kindling,guo2020zero}, \textbf{JED}-based methods (Jointly optimization Enhancement and Detection)\cite{sun2022rethinking,yin2023pe,qin2022denet,liu2022image,kalwar2023gdip,ogino2025erup,feng2024toward,feng2025lightstar,ji2025fcma}, 
    and \textbf{IIFE}-based methods (Illumination-Invariant Feature Extraction)\cite{cui2021multitask,kennerley20232pcnet,zhang2024isp,du2024boosting,hong2024you}.
    Among them, As shown in Fig. \ref{f1} (b), LLIE-based methods can improve image brightness before inputting a detector trained on well-lit images, but the enhanced images may not suit for detector. Fig. \ref{f1} (c) illustrates the JED-based methods that jointly optimize enhancement and detection networks during the training process. These kind of methods enable the enhance images to better suit the detection task. IIFE-based methods learn illumination-invariant features by adopting the strategies like multi-task learning, domain adaptation, color channel transform, and so on, thereby enhancing detection performance on dark images.

    Although some effective optimization strategies exist for low-illumination object detection, most rely on a single improvement scheme. 
    However, in practical application, image characteristics vary significantly across different scenes and imaging conditions, making it difficult for a single improvement scheme to adapt flexibly to diverse variations.
    Therefore, inspired by unsupervised label distribution method \cite{9873970}, we propose an Adaptive  Multi-Experts Image Enhancement Object Detection (AMIEOD) framework by introducing two detector-compatible preprocessing modules. The main innovations can be summarized as follows.
    \begin{itemize}
        \item To ensure the detector can fully acquire features of low-illumination images from various perspectives, we introduce a Multi-Experts Image Enhancement Module (MEIEM) for preprocessing, which integrates multiple enhancement models optimized with diverse strategies.
        \item To enable the sub-modules of MEIEM to collaboratively learn information beneficial for detection, we propose Detection-Guided Regression Loss (DGRL), which optimizes MEIEM based on the detection results of its brightness-enhanced outputs.
        \item We construct an Expert Selection Module (ESM) and Detection-Guided Cross-Entropy (DGCE) loss to adaptively select an appropriate enhancement strategy from MEIEM according to image characters.		
	\end{itemize}

\section{Related Work \label{RelatedWork}}

\subsection{Object Detection}

    Object detection is a crucial task in computer vision, applied in areas like facial recognition\cite{11086396,9387154}, autonomous driving \cite{11071878}, and security monitoring \cite{li2023yolosr}. With the rise of deep learning, two main approaches have emerged: 1) region-based and 2) regression-based methods.
    \par 
    Region-based methods, such as R-CNN\cite{girshick2014rich}, generate proposals and then classify them using models like support vector machine(SVM). More variants like Fast R-CNN\cite{girshick2015fast}, Faster R-CNN\cite{ren2015faster} and Mask R-CNN \cite{he2017mask}, among others. have further developed detection algorithm but tend to be slower due to their two-stage process. 
    \par 
    On the other hand, regression-based methods like YOLO\cite{zhang2023superyolo,huang2025crkd,zhang2025frfcnet}, SSD\cite{liu2016ssd}, and RetinaNet\cite{lin2017focal} predict object categories and locations in a single step. They use preset anchor boxes to speed up the process and are optimized by techniques like Focal Loss\cite{lin2017focal} and Feature Pyramid Networks\cite{lin2017feature}, making them faster and more efficient, though slightly less accurate than region-based methods.
    Recently, transformer\cite{vaswani2017attention} based models like DETR\cite{carion2020end} have become popular, replacing traditional convolutional approaches by treating detection as a direct set prediction problem, removing the need for region proposals or anchor boxes. Variants like RT-DETR \cite{zhao2024detrs} and DINO\cite{zhang2022dino} have improved accuracy and speed with advanced attention mechanisms and multi-scale features.
    With the assistance of vision-language model like CLIP\cite{radford2021learning}, Zero-shot detection\cite{cao2025survey} and Open-vocabulary detection \cite{10948367} make great progress, such as Grounding DINO\cite{10768306}, can recognize unseen object categories, leveraging pre-trained language models alongside visual transformers for better generalization.
    \par 
    In summary, regression-based methods excel in real-time performance, while region-based methods offer better accuracy at the cost of speed. The choice between them depends on the trade-off between accuracy and speed for specific applications.
    

\subsection{Object Detection on Low-illumination Scenes}

	Although object detectors have advanced significantly, their performance in low-illumination scenes remains limited. While some night-time methods exploit infrared images to improve robustness\cite{10478590}, they require additional sensors that are often unavailable in practice. Therefore, enabling real-time and efficient night-time detection using only a single visible input remains an open challenge.
    To address this issue, an intuitive approach is to utilize pretrained image enhancement models, such as MBLLEN \cite{lv2018mbllen}, KIND \cite{zhang2019kindling}, Zero-DCE \cite{guo2020zero}, and SCI \cite{ma2022toward}, as preprocessing modules to improve image brightness before feeding the enhanced images into a detector trained on well-lit data. 
    While these two-stage approaches intuitively enhance image clarity and make object features more prominent, an inevitable domain gap persists between the enhanced images and the well-lit images expected by detection models. This discrepancy continues to hinder the effectiveness of visual perception in low-illumination scenarios.
    \par 
    To enable enhanced images to better align with detection tasks, several JED-based methods jointly optimize enhancement and detection networks during the training process. For instance, Sun et al. \cite{sun2022rethinking} propose an adversarial learning paradigm to generate enhanced images that are more suitable for detection models. PEYOLO \cite{yin2023pe} and DEYOLO \cite{qin2022denet} introduce a pyramid enhancement network based on the Laplacian pyramid, specifically designed for object detection in low-illumination scenes.
    Some studies further explore image-adaptive enhancement methods through joint optimization with detection. For example, IAYOLO 
    \cite{liu2022image}, GDIP \cite{kalwar2023gdip}, and ERUP-YOLO \cite{ogino2025erup} incorporate differentiable image processing (DIP) and trainable parameter prediction (PP) modules to adaptively adjust image contrast, white balance, tone, etc., thereby improving detection performance under adverse environmental conditions.
	EMV-YOLO\cite{feng2024toward} dynamically adapts to the object detection based on end-to-end training and emphasize the semantic information for the detection.
	While these JED methods enhance the suitability of images for detection tasks, they fall short in fully capturing the information distribution from dark images \cite{cui2024trash}. Subsequent methods have therefore explored improvements from this perspective.
	LightStar-YOLO\cite{feng2025lightstar} uses inverse mapping to convert RGB images into Pseudo-RAW feature space to retain more information.
	FCMA-Det\cite{ji2025fcma} performs dynamic fusion of the original image and the enhanced image information for the low-illuminatino detection algorithm.
	BAD-Net\cite{10012056} designs a detection-oriented enhance framework to reduce enhancement–detection mismatch and improve object detection in real adverse weather scenes.

    \par 
    Additionally, IIFE-based algorithms employ various strategies to learn illumination-invariant features, thereby enhancing detection performance on dark images. Specifically, MAET \cite{cui2021multitask} improves robustness in dark object detection through a multi-task learning approach, while 2PCNet \cite{kennerley20232pcnet}, ISP-teacher \cite{zhang2024isp}, and DAINet \cite{du2024boosting} leverage domain adaptation from day-to-night data.
    Although these methods can partially improve robustness to illumination variations, they heavily rely on day-to-night data modeling (e.g., gamma correction, white balance, and sharpening). However, these parameters are difficult to maintain consistently across different imaging devices and environments, which continues to limit the generalization capability of detection algorithms. 
    YOLA \cite{hong2024you} learns illumination-invariant features by incorporating a preprocessing module and illumination-invariant loss, while Cui et al. \cite{cui2024trash} separate image "content" and "illumination" components based on Retinex theory \cite{land1971lightness}, then re-integrate them at the feature level.
    However, both approaches struggle to acquire high-quality representations from well-lit images during the training phase.
    \par 
    

\section{Methodology}
    
    \begin{figure*}[ht]
		\centering
		\includegraphics[scale=0.58]{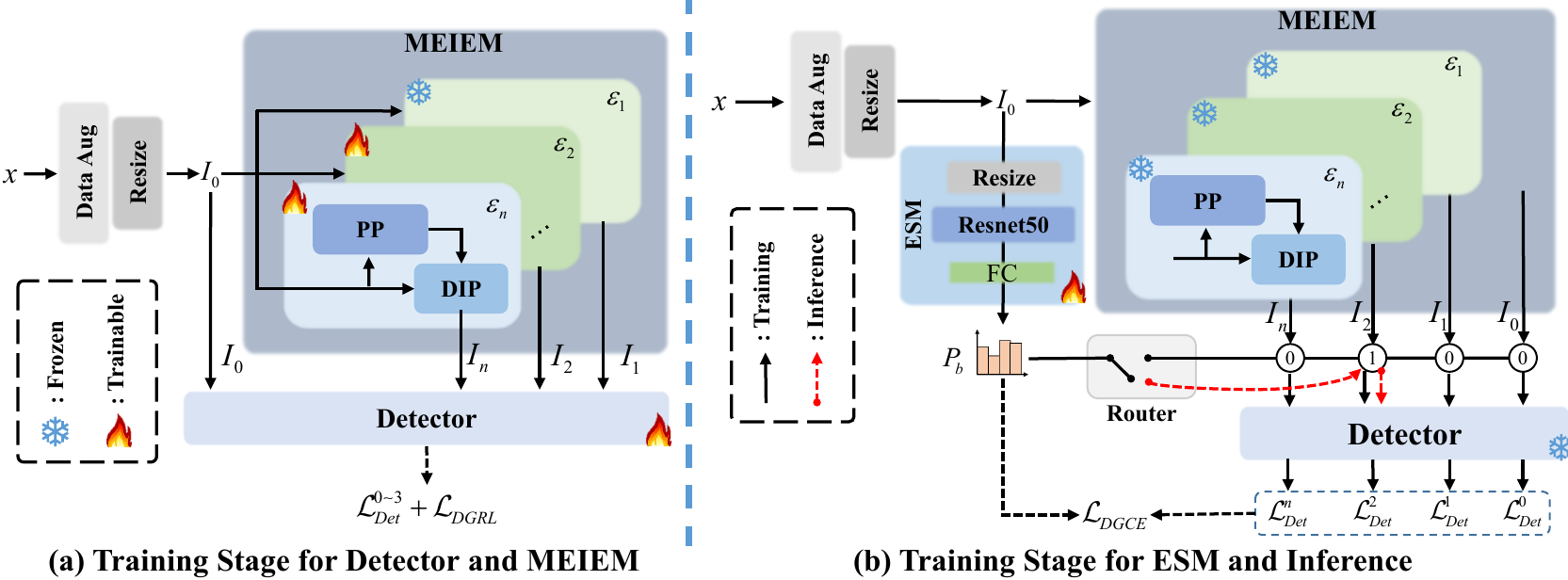}
		\caption{Overview of our proposed method. ``FC'' is the full connection layer. The training process consists of two stages. The first stage trains MEIEM and detector. The second stage trains ESM for expert selection. During inference, the ESM outputs confidence scores to guide a routing mechanism that selects an appropriate preprocessing module from MEIEM, followed by enhancement and object detection.}
		\label{f2}
    \end{figure*}

	In this section, we first present an overview of AMIEOD, followed by detailed descriptions of MEIEM, ESM, and the associated DGRL and DGCE.

	\subsection{Overview of AMIEOD}
	
	The proposed AMIEOD framework incorporates a MEIEM to comprehensively explore low-illumination image characteristics from multiple perspectives, thereby enhancing the detector’s capability to learn and recognize objects in poorly lit scenarios. In addition, an ESM is introduced to adaptively select the most suitable enhancement strategy from MEIEM based on the characteristics of the input image, further improving the flexibility and performance of the detection system.
	\par
	\subsubsection{Training Process}
	
	As illustrated in Fig.~\ref{f2}, the training procedure is divided into two stages. In the first stage, MEIEM and the detection network are jointly optimized in an end-to-end manner using the detection loss and the proposed DGRL. During this stage, multiple enhanced images together with the original input emphasize different aspects of image information, and jointly serving as inputs provides richer and more diverse visual representations for the detector.
	\par
	
		In the second training stage, the enhancement experts and the detector pretrained in the first stage are fixed, and an ESM is trained using the proposed DGCE loss. This stage enables the ESM to predict the most suitable enhancement strategy for each input image according to its global characteristics.
	
	\subsubsection{Inference Process}
	
		During inference, as shown in Fig.~\ref{f1}(d) and Fig.~\ref{f2}(b), a given input image $I_0$, after data augmentation and resizing, is first processed by the ESM to output confidence scores for all enhancement strategies. Based on these scores and a gating mechanism, the optimal enhancement strategy is selected and applied to the input image, which is then fed into the detector to obtain the final detection result. This process can be formulated as:
	\begin{equation}\label{q1}
		P_b = \arg\max(\mathrm{softmax}(f^{esm}(I_0))) \in \{0,1,2,\dots,n\}
	\end{equation}
	\begin{equation}\label{q2}
		P(I_0) = f^{Det}\!\left(f^M_{k=P_b}(I_0)\right)
	\end{equation}
	where $P_b$ denotes the enhancement strategy selected by the ESM, $f^{esm}(\cdot)$ represents the ESM function, $f^{Det}(\cdot)$ denotes the detection network, $f^M_k(\cdot)$ is the $k$-th enhancement expert in MEIEM, and $P(I_0)$ is the final detection result for $I_0$.


	%
\subsection{Multi-Experts Image Enhancement Module (MEIEM)}	

    As illustrated in Fig. \ref{f2}, MEIEM integrates multiple enhancement modules to improve the input image from various perspectives.  In the first training stage, MEIEM employs diverse image preprocessing strategies to extract complementary information, enabling the detection network to more effectively learn and leverage the informative content of dark images.
    In the second stage, the various enhanced images generated by MEIEM are fed into the network for detection, and their corresponding loss values are computed. These loss values are then used to train the ESM to select the most suitable enhancement strategy from MEIEM. During inference, the ESM dynamically selects the most appropriate enhancement strategy for each input image.
    \par 
    \par 
    Specifically, for a given input image $I_0$, it is fed separately into three sub-modules of MEIEM, producing different enhanced images $I_1$, $I_2$, ..., $I_n$. This process can be defined as follows:
    \begin{equation}\label{q3}
	I_k = f^M_k(I_0), k \in 1, 2, ..., n
    \end{equation}
	We further introduce DGRL to select the one with the lowest detection loss among $I_{0 \sim n}$ as the target for calculating regress loss. This guides the parameter update for MEIEM via backpropagation, thereby enhancing detection performance.
	Above all, the overall loss for MEIEM and detector in this stage can be formulated as follow:
	\begin{equation}\label{q4}
    	\mathcal{L}_{stage1}=(1-\alpha)\mathcal{L}_{DGRL} + \frac{\alpha}{n+1} \sum_{k=0}^{n} \mathcal{L}_{Det}(I_k, GT)
	\end{equation}
	where $\mathcal{L}_{Det}(\cdot, \cdot) $ expresses the current detection loss, which consists of   $\mathcal{L}_{box}$, $\mathcal{L}_{obj}$, $\mathcal{L}_{cls}$. Among them, 
	$ \mathcal{L}_{box} $ measures the discrepancy between the predicted and ground-truth bounding boxes and is computed using the CIoU loss. The objectness loss $\mathcal{L}_{obj}$ and classification loss $\mathcal{L}_{cls}$ are implemented using binary cross-entropy loss for bounding box confidence and class prediction, respectively.
	GT is the label of $I_0$. $\mathcal{L}_{DGRL}$ indicates the DGRL and it will be concretely introduced in the subsequent section. $\alpha$ is the coefficients between DGRL and detection loss, which is set to 0.2 in our method.

	\begin{figure}[h]
		\centering
		\includegraphics[scale=0.65]{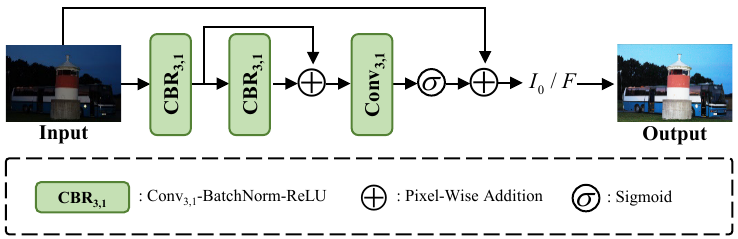}
		\caption{The model structure of PIEM and JIEM.}
		\label{SCI}
	\end{figure}

	In MEIEM, we incorporate a Pretrained Image Enhancement Module (PIEM), a Jointly Optimized Enhancement Module (JIEM), and an Image-Adaptive Enhancement Module (IAEM). PIEM adopts the pretrained SCI (Self-Calibrated Illumination) model \cite{ma2022toward}, whose parameters are frozen to enhance image brightness and provide clearer object cues. 
	JIEM shares the same architecture as PIEM, but its parameters are jointly optimized with the detector using the detection loss and the proposed DGRL, enabling image brightness enhancement while ensuring consistency with detection-aware representations.
	As illustrated in Fig.~\ref{SCI}, both PIEM and JIEM follow a lightweight structure consisting of two cascaded CBR (Convolution–BatchNorm–ReLU) blocks, whose outputs are summed and fed into a convolutional layer with three output channels. After sigmoid activation, the resulting features are fused with the input image and used to normalize it, producing the enhanced output.

	\par 
	\begin{figure}[h]
		\centering
		\includegraphics[scale=0.46]{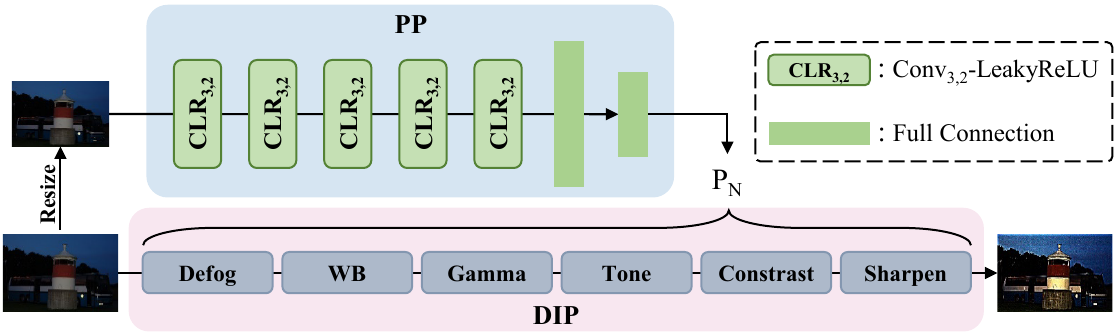}
		\caption{The structure of IAEM. ``PP'' is the parameter prediction module. ``DIP'' indicates a differentiable image processing.}
		\label{IAE}
	\end{figure}
	The IAEM in MEIEM adopts the image-adaptive enhancement strategy proposed in IAYOLO~\cite{liu2022image}. As illustrated in Fig.~\ref{IAE}, it consists of a Parameter Prediction (PP) module and a Differentiable Image Processing (DIP) module.
	The PP module is designed to predict image-adaptive parameters for the DIP module based on global image characteristics (e.g., brightness, color, and tone). Each input image is first resized to 256 × 256 via bilinear interpolation and then fed into the PP network, which comprises five CLR (Convolution–LeakyReLU) blocks followed by two fully connected layers. The convolutional layers use channel sizes of 16, 32, 32, 32, and 32, while the two fully connected layers output 64 and 15 dimensions, respectively. The PP module produces a parameter vector $P_N$ $\in \mathcal{R}^{1 \times 15}$, which is used to configure the DIP module.

    \subsection{Detection-Guided Regression Loss (DGRL)  }	
	When training MEIEM, the detection performance for $I_{0 \sim n}$ varies. Ensuring that the sub-module in MEIEM learns from the target with the best detection performance among $I_{0 \sim n}$ can achieve more excellent performance. 
	This can be interpreted as an approximation of empirical risk minimization under the detection task. The computation of DGRL is summarized in Algorithm~\ref{DGRL A}. Let $f_k^M(I_0)$ denote the enhanced image produced by the $k$-th expert. The enhancement output that best aligns with the current detector is defined as
	\begin{equation}\label{q5}
		b =\arg\underset{k}{\min}(\mathcal{L}_{Det}(f_k^M(I_0), GT))
	\end{equation}
	which implicitly defines a task-oriented target for enhancement optimization. DGRL enforces consistency among experts by regressing other enhanced outputs toward $I_b$, while gradient truncation via $detach()$ prevents updating the parameters of the selected expert. Formally, DGRL minimizes
	\begin{equation}\label{q6}
		\mathcal{L}_{DGRL} = \frac{1}{n} \sum_{k=0}^{n} \left \| I_{k}-I_b\text{.detach()} \right \| _{1}   
	\end{equation}
	which corresponds to reducing the variance of multi-expert outputs around the optimal enhancement solution.
	\par 
	In $L_{\text{DGRL}}$, since regression losses are computed only among MEIEM outputs, gradients are confined to the enhancement modules, avoiding interference with detector optimization. 
	Moreover, the multi-expert design allows the regression target to be selected from an expert with lower detection loss when others produce noisy outputs, thereby suppressing noise during early training.

	\begin{algorithm}
		\caption{DGRL Algorithm}
		\label{DGRL A}
		\begin{algorithmic}[1]
			\STATE \textbf{Input:} \(I_0\) 
			\STATE \(I_k \leftarrow \text{Eq. (3)}\), \(b \leftarrow \text{Eq. (5)}\)
			\FOR{$k = 0 \text{ to } n$}
			\STATE \(\mathcal{L}_{\text{DGRL}} \leftarrow \sum_{k=0}^{n}|I_k - I_b.\text{detach()}|\)
			\ENDFOR
			\RETURN \(\mathcal{L}_{\text{DGRL}}\)
		\end{algorithmic}
	\end{algorithm}

    
	
	\subsection{Expert Selection Module (ESM)}	
	During the first training stage, the MEIEM employs multiple enhancement strategies to capture image information from diverse perspectives, boosting detection performance. 
	During inference, only one preprocessing scheme of MEIEM can be used, limiting adaptability to diverse scene conditions. 
	Dynamically selecting an appropriate enhancement based on input characteristics remains a key challenge. To address this, we introduce the ESM, which selects the enhancement strategy suited for the input image by analyzing the global characteristics of the input image. 
	\par 	
	As shown in Fig. \ref{f2}, ESM consists of a ResNet50 and a fully connection layer. 
	The input image is first resized to reduce computational overhead while retain globe information. 
	Then, the image is processed through ESM, which outputs the confidence scores corresponding to the likelihood of $I_{0 \sim n}$ yields the best detection performance when fed into the detection network. The above process can be detailed as 
	\begin{equation}\label{q7}
		f^{esm}(I_0)=FC(Res(resize(I_0)))
	\end{equation}
	where $FC(\cdot)$ and $ResNet(\cdot)$ refer to the function of full connection layer and ResNet. $resize(\cdot)$ expresses the downsample of input.
	Since the ESM is optimized with the proposed DGCE loss, it requires an additional training stage using the pretrained MEIEM and detection network. 
	We will introduce the optimization process of ESM in the following section in detail.

	\subsection{Detection-Guided Cross Entropy (DGCE)}
	
	The DGCE loss can be theoretically interpreted as a maximum-likelihood estimation problem under the assumption that the optimal expert corresponds to the one minimizing detection loss. It should be noted that DGCE is designed to optimize the ESM for selecting, rather than learning, the enhancement strategy from the already-trained MEIEM that is most compatible with the current detector.
	Therefore, we divide the optimization process of the ESM into three steps:\\
	\textbf{(1) Ground-Truth Label Assignment:} for each input image, we evaluate the detection performance of the original and three enhanced variants using the pretrained detection network. The enhancement strategy $b$ that yields the minimum detection loss is selected as the ground-truth label.\\
	\textbf{(2) Confidence Score Prediction:} the input image is downsampled and passed through ResNet-50, followed by a fully connected layer that outputs a four-dimensional score vector. Each score corresponds to the likelihood that a given enhancement strategy is optimal for the input.\\
	\textbf{(3) Loss Computation and Optimization:} A cross-entropy loss is computed between the predicted scores and the assigned ground-truth label. This loss is used to optimize the ESM, enabling it to predict the most suitable enhancement strategy based on image characteristics.
	Thus, the expected detection loss can be reduced even further by the experts selection.
	\begin{equation}\label{q7}
		min\mathbb{E}_{I_0}[\mathcal{L}_{Det}(f^M_{k=P_b}(I_0), GT)]
	\end{equation}
	\par 
	For any input sample, DGCE optimizes the posterior probability
	\begin{equation}\label{q7}
		p(k|I_0) = f^{esm}(I_0),
	\end{equation}
	by minimizing
	\begin{equation}\label{q8}
		\mathcal{L}_{DGCE} = -\text{log} p(b|I_0)
	\end{equation}
	where $b$ is the same calculation process as Eq. \ref{q5}, corresponds to the enhancement strategy within MEIEM that minimizes detection risk under the current detector. $k$ denotes the predicted enhancement expert that yields the lowest detection loss. This defines an implicit optimal decision rule conditioned on $I_0$.
	\par
	From a probabilistic perspective, the ESM aims to approximate the posterior distribution $p(k|I_0)$. Training ESM with DGCE is equivalent to maximizing the log-likelihood of the optimal expert index $b$ under a categorical distribution parameterized by ESM. Minimizing the loss in Eq. \ref{q8} thus corresponds to maximum likelihood estimation of the expert-selection policy. 
	\par 
	Importantly, DGCE leverages task-level supervision derived from detection performance rather than low-level enhancement metrics. Moreover, since the detector is frozen during DGCE optimization, the pseudo-label assignment remains stable and will not cause the accumulation of learning noise or model collapse between ESM and detector. This design prevents error amplification and ensures that the ESM converges toward selecting the enhancement strategy that minimizes expected detection risk.
	Therefore, DGCE provides a principled learning objective that aligns expert selection with the ultimate detection task, offering a theoretically grounded alternative to heuristic routing or rule-based selection mechanisms. 
	
	\par
	\subsection{Differences with Prior Works }

	In recent years, several researches have explored image enhancement techniques to improve object detection under low-illumination conditions. This section summarizes the differences between our AMIEOD and representative methods, including PEYOLO \cite{yin2023pe}, IAYOLO \cite{liu2022image}, GDIP \cite{kalwar2023gdip}, MAET \cite{cui2021multitask}, and YOLA \cite{hong2024you}.
	PEYOLO \cite{yin2023pe} is a typical JED-based approach that jointly optimizes a learnable Laplacian pyramid network with the detector. IAYOLO \cite{liu2022image} and GDIP \cite{kalwar2023gdip} adopt image-adaptive enhancement strategies, while YOLA \cite{hong2024you} focuses on learning illumination-invariant features.
	\par 
	\subsubsection{Differences with PEYOLO\cite{yin2023pe}}
	PEYOLO is a typical JED-based method that jointly optimizes a learnable Laplacian pyramid enhancement network with the detector. In contrast, AMIEOD introduces a MEIEM that integrates multiple heterogeneous enhancement strategies, together with an adaptive expert selection mechanism to dynamically choose the most suitable enhancement during inference, enabling greater flexibility under diverse low-illumination conditions.
	\subsubsection{Differences with IAYOLO~\cite{liu2022image} and GDIP~\cite{kalwar2023gdip}}
	IAYOLO adopts an image-adaptive enhancement strategy based on differentiable image-filtering, where filtering parameters are predicted by a parameter prediction network for each input image, while GDIP improves robustness under adverse conditions by embedding gated differentiable image-filtering operations into the detection framework.
	\par 
	
	In contrast, AMIEOD models illumination diversity by introducing multiple enhancement experts, including the image-adaptive module from IAYOLO, which are jointly optimized with the detector in a detection-guided manner using the proposed DGRL. Based on the trained multi-enhancement experts and detector, an ESM is further trained with the proposed DGCE loss to adaptively select the most suitable enhancement strategy for each input image during inference according to its image characteristics.
	\subsubsection{Differences with MAET\cite{cui2021multitask}}

	MAET improves object detection in poorly lit scenes by learning illumination-invariant features through multi-task learning, whereas AMIEOD focuses on detection-oriented image enhancement. By jointly optimizing enhancement and detection using detection-guided losses, AMIEOD directly aligns the enhancement process with detection objectives.
	\subsubsection{Differences with YOLA\cite{hong2024you}}
	
	YOLA learns illumination-invariant representations through a specialized preprocessing module, whereas AMIEOD adopts a multi-expert enhancement framework with adaptive expert selection. Instead of enforcing a unified illumination-invariant representation, AMIEOD dynamically selects the most suitable enhancement strategy for each input image, providing greater flexibility in diverse low-illumination environments.

	\par 

	\section{Experiments}
	%
	
	We conduct comprehensive experiments on Exdark \cite{loh2019getting}, and comparative studies on Darkface \cite{yang2020advancing}, LLVIP \cite{jia2021llvip}, and M3FD \cite{liu2022target} to evaluate the generalization of our method. \textbf{All experiments use YOLOv3 \cite{redmon2018yolov3} as the baseline unless otherwise specified.}

	\begin{table}[h]
		\caption{Details of the adopted low-illumination datasets. ``NoC.'' indicates the Number of Category. ``Mod.'' represses the modality of dataset.}
		\centering 
		\small
		\renewcommand{\arraystretch}{1}
		\setlength{\tabcolsep}{2.8mm}{
			\begin{threeparttable}
				\begin{tabular}{cccccc}  
					\toprule
					Dataset&Train&Test&Total&NoC.&Mod.\\
					\midrule
					Exdark\cite{loh2019getting} & 5896&1467&7363&12&Vis	\\
					
					Darkface\cite{yang2020advancing}&5400&600&6000&1 &Vis	\\
					LLVIP\cite{jia2021llvip}&12025&3463&15488&1 &Vis-IR
					\\
					M3FD\cite{liu2022target}&3500&700&4200&6 &Vis-IR\\
					
					\bottomrule
					
				\end{tabular}
		\end{threeparttable}	}
		\label{T1} 	
	\end{table}
	
	\subsection{Datasets}
	
	Table \ref{T1} lists the details of the datasets used. The images from four datasets are primarily captured under low-illumination conditions. 
	The data partitioning for Exdark follows MAET \cite{cui2021multitask}, while M3FD is randomly split. Note that LLVIP and M3FD are Visible-InfraRed datasets, but we only use the visible images in our experiments.

	\begin{table*}[ht]
		\centering 
		\renewcommand{\arraystretch}{1}
		\setlength{\tabcolsep}{1.6mm}{
		\caption{Quantitative detection results on the ExDark. The best result is highlighted in bold. `*'' means the enhancement and detection models are jointly trained. We report the AP for all categories, from left to right: Bicycle, Boat, Bottle, Bus, Car, Cat, Chair, Cup, Dog, Motorbike, People, Table. ``P'' refers to precision rate. ``R'' indicates recall rate. Best results are highlighted in \textbf{bold}.}
		\label{Compared table} 
			\begin{threeparttable}
				\begin{tabular}{c|c|cccccccccccc|ccc}  
					\toprule
					Method &Type  &Bic.&Boa.&Bot.&Bus&Car&Cat&Cha.&Cup&Dog&Mot.&Peo.&Tab.&P&R& mAP(\%)		\\
					\midrule
					
					YOLOv3\textsuperscript{[arXiv,2018]} &- &83.1&75.5&79.7&91.9&84.4&68.9&68.2&75.1&75.4&77.4&80.2&58.3
					&82.3&70.3&76.5\\ 				
					\midrule
					KIND-YOLOv3\textsuperscript{[ACM MM,2019]}  & \multirow{4}*{LLIE}  &79.4&74.6&77.0&92.7&84.2&71.1&66.5&73.4&78.2&75.4&79.6&55.5
					&84.3&67.7&75.7 \textcolor{green}{(-0.8)}\\ 
					MBLLEN-YOLOv3 \textsuperscript{[BMVC,2018]}  &    &82.6&77.1&81.0&92.5&83.9&71.5&68.4&74.3&76.9&78.2&80.1&57.7
					&84.4&68.9&77.0 \textcolor{blue}{(+0.5)}  \\  
					ZeroDCE-YOLOv3\textsuperscript{[CVPR,2020]} & &82.9&73.1&78.9&91.0&85.3&76.1&66.9&75.1&79.9&77.6&80.5&57.3
					&\textbf{85.6}&68.6& 77.1 \textcolor{blue}{(+0.6)}   \\ 	
					SCI-YOLOv3\textsuperscript{[CVPR,2022]} &&82.7&74.6&77.5&91.0&84.1&73.0&67.8&75.0&79.6&77.1&79.3&57.1
					&84.2&70.7&  76.6 \textcolor{blue}{(+0.1)} \\

					\midrule
					SCI-YOLOv3* \textsuperscript{[CVPR,2022]}&\multirow{7}*{JED}&83.9&75.1&79.4&90.8&85.2&75.0&69.6&76.6&80.8&76.0&80.5&57.2&84.2&70.7&77.4 \textcolor{blue}{(+0.9)}\\
					IAYOLO\textsuperscript{[AAAI,2022]}  &&83.4&77.4&80.4&90.7&84.5&73.7&68.9&75.4&80.7&79.4&81.1&57.0
					&82.9&71.3& 77.8 \textcolor{blue}{(+1.3)}\\ 
					 
					PEYOLO \textsuperscript{[ICANN,2023]} &&\textbf{84.7}&79.2&79.3&92.5&83.9&71.5&71.7&79.7&79.7&77.3&81.8&55.3&-&-&78.0 \textcolor{blue}{(+1.5)}  \\
					DEYOLO \textsuperscript{[ACCV,2022]}&&80.4&79.7&77.9&91.2&82.7&72.8&69.9&80.1&77.2&76.7&82.0&57.2
					&-&-& 77.3 \textcolor{blue}{(+0.8)}\\

					EMV-YOLO\textsuperscript{[BMVC,2024]}&&82.8&79.7&79.8&94.1&84.7&74.3&74.1&83.1&82.7&78.1&83.6&59.3&-&-&79.7\textcolor{blue}{(+3.2)}\\
					LightStar-YOLO\textsuperscript{[ICCVM,2025]}&&83.6&80.5&77.0&91.4&84.8&73.9&73.9&80.6&78.1&78.6&82.1&58.4&-&-&78.6\textcolor{blue}{(+2.1)}\\
					FCMA-Det\textsuperscript{[TGRS,2025]}&&83.9&80.2&79.4&92.8&84.5&74.8&73.6&81.2&81.1&\textbf{80.7}&83.4&59.2&-&-&79.6\textcolor{blue}{(+3.1)}\\
					\midrule
					
					MAET\textsuperscript{[ICCV,2021]}&\multirow{3}*{IIFE}&83.1&78.5&75.6&92.9&73.1&73.1&71.3&79.9&79.8&77.2&81.1&57.0&-&-&77.7 \textcolor{blue}{(+1.1)}
					\\
					DAINet\textsuperscript{[CVPR,2024]}&&83.8&75.8&75.1&94.2&84.1&74.9&73.1&78.2&82.2&76.4&80.7&59.8&-&-&78.3 \textcolor{blue}{(+1.8)}
					\\
					YOLA\textsuperscript{[NeurIPS,2024]}&&83.0&76.5&79.9&93.0&85.3&\textbf{76.3}&69.2&77.0&81.0&79.5&80.9&59.1&82.6&72.8&78.4 \textcolor{blue}{(+1.9)}
					\\
					\midrule
					
                    \rowcolor {gray!20}
                    \textbf {AMIEOD (Our)}&-&83.5&\textbf {81.6}&\textbf {83.2}&\textbf {95.5}&\textbf {88.0}&75.9&\textbf {76.7}&\textbf
                    {85.8}&\textbf {84.7}&80.2&\textbf {86.0}&\textbf {64.1}&79.4&\textbf {76.3}&\textbf {82.1 \textcolor{blue}{(+5.6)}} \\

					\bottomrule
				\end{tabular}	
		\end{threeparttable}		}
	\end{table*}

\subsection{Implementation Details}

\subsubsection{Experimental Setting}
	There are two training stages for our AMIEOD. \textbf{In the first stage}, the model is optimized based on the YOLOv3 network pretrained on the COCO dataset, with the batch size set to 8, initial learning rate set to 0.01, and the training epochs set to 100. \textbf{In the second stage}, the initial learning rate is reduced to 0.001, and the epoch is set to 30, with a batch size of 1 to assign the optimal strategy for a single input. 
    \par 
    The SGD \cite{bottou2012stochastic} optimizer is employed, with a momentum coefficient of 0.937 and a weight decay of 0.0005. All input images are resized to a fixed size of 640 $\times$640. 
	We employ the widely used precision, recall, and mean Average Precision with the IOU threshold set to 50\% 
    (mAP$_{50}$) as the evaluation metric for detection performance. The experiments are conducted on GeForce RTX 3090 GPUs.

\subsection{Comparisons with Current Methods}
	
\subsubsection{Comparison Methods}
	There are four LLIE methods, including KIND \cite{zhang2019kindling}, MBLLEN \cite{lv2018mbllen}, ZeroDCE \cite{guo2020zero}, and SCI \cite{ma2022toward};  three JED methods, including IAYOLO\cite{liu2022image}, PE-YOLO \cite{yin2023pe}, 
	DE-YOLO \cite{qin2022denet}, EMV-YOLO\cite{feng2024toward}, LightStar-YOLO\cite{feng2025lightstar}, and FCMA-Det\cite{ji2025fcma};
	and three IIFE methods, including MAET \cite{cui2021multitask}, DAINet \cite{du2024boosting}, and YOLA \cite{hong2024you} are compared. 
	Moreover, since our proposed MEIEM integrates the SCI enhancement network via joint optimization, we also compare the method in which the SCI\cite{ma2022toward} network is jointly optimized with the detection model.
	\par

    \begin{figure*}[ht]
    	\centering
    	\includegraphics[scale=0.37]{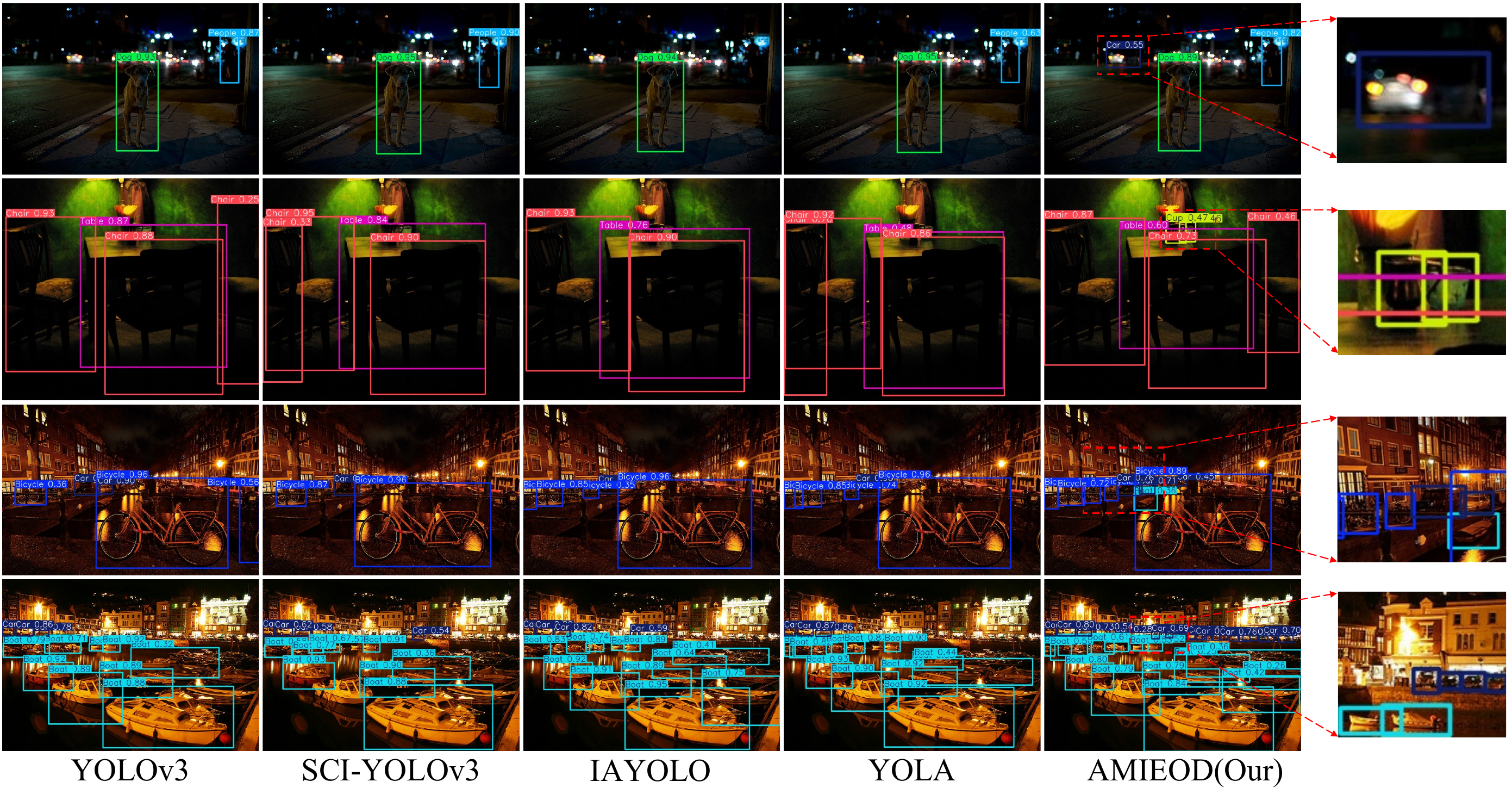}
    	\caption{Visualization results on Exdark dataset. The areas outlined by red dashed lines are enlarged for better visualization. }
    	\label{f4}
    \end{figure*}
\subsubsection{Quantitative Results and Analysis}
	Table \ref{Compared table} summarizes the experimental results of various low-illumination object detection methods. Overall, LLIE-based methods provide limited improvement for detection performance. For example, the best-performing LLIE method, Zero-DCE, improves mAP by only 0.6\%, while KIND even degrades performance by 0.8\%, and MBLLEN yields a marginal gain of 0.5\%.
	\par 
	JED-based methods consistently outperform LLIE-based approaches, indicating that task-aware enhancement is more effective than independent image preprocessing. Specifically, IAYOLO, PE-YOLO, and DE-YOLO achieve mAP scores of 77.8\%, 78.0\%, and 77.3\%, corresponding to improvements of 1.3\%, 1.5\%, and 0.8\% over the baseline, respectively. Moreover, three recent JED-based methods—EMV-YOLO, LightStar-YOLO, and FCMA-Det—achieve mAP values of 79.7\%, 78.6\%, and 79.6\%, respectively.
	\par 
	Several methods also adopt IIFE-based strategies, among which MAET, DAINet, and YOLA achieve mAP scores of 77.7\%, 78.3\%, and 78.4\%, respectively.
	\par
	
	In contrast, the proposed AMIEOD achieves an mAP of 82.1\% on the ExDark dataset, outperforming the baseline detector by 5.6\% and surpassing all compared methods across different enhancement paradigms. Notably, AMIEOD improves mAP by 2.4\% over the current state-of-the-art EMV-YOLO (79.7\% vs. 82.1\%), demonstrating its superior effectiveness for low-illumination object detection.
	
	\par 
	Our method also achieves a significantly higher recall compared to existing approaches. While it shows lower precision, this is likely due to many correctly detected targets being unannotated in ground truth.
	\par 
	Moreover, AMIEOD achieves the highest AP in 9 out of 12 object categories, only with a slight drop by 1.2$\%$ (84.7$\%$ vs. 83.5$\%$) in ``bicycle'' compared to PEYOLO and 0.4$\%$ (76.3$\%$ vs. 75.9$\%$) in ``Cat'' compared to YOLA, as well as 0.5\% in ``Motorbike'' compared to FCMA-Det. This demonstrates the stability and robustness of the proposed approach across diverse object classes in low-illumination scenarios. It is worth noting that the AP for the ``Table'' category is consistently lower than that of other categories across all algorithms listed in Table \ref{Compared table}. This is primarily due to the scarcity of training samples for the ``Table'' in the ExDark dataset.



    		
	
	\subsubsection{Qualitative Results and Analysis}

	Fig. \ref{f4} presents a visual comparison of detection results between the proposed AMIEOD method and representative algorithms from the LLIE, JED, and IIFE categories, specifically SCI-YOLOv3, IAYOLO, and YOLA, as well as the baseline YOLOv3. 
        \par 
	Both the JED-based IAYOLO and the IIFE-based YOLA algorithms demonstrate enhanced detection capabilities compared to the baseline, exhibiting a stronger focus on target regions. However, despite these improvements, the detection accuracy for certain targets in complex background areas remains insufficient. This suggests that although the integration of joint enhancement or feature enhancement modules contributes to overall performance gains, there is still significant room for further optimization.
        \par
	In contrast, the proposed AMIEOD consistently achieves superior detection performance across a wide range of low-illumination scenarios. Our approach demonstrates a strong ability to accurately detect a greater number of targets, particularly those that are small, dimly lit, or partially occluded. This highlights the effectiveness of our method in preserving essential structural and semantic information, thereby significantly enhancing detection robustness under challenging illumination conditions.
    


	
	
\subsubsection{Heatmap Visualization Results} 
		
	\begin{figure*}[ht]
		\centering
		\includegraphics[scale=0.73]{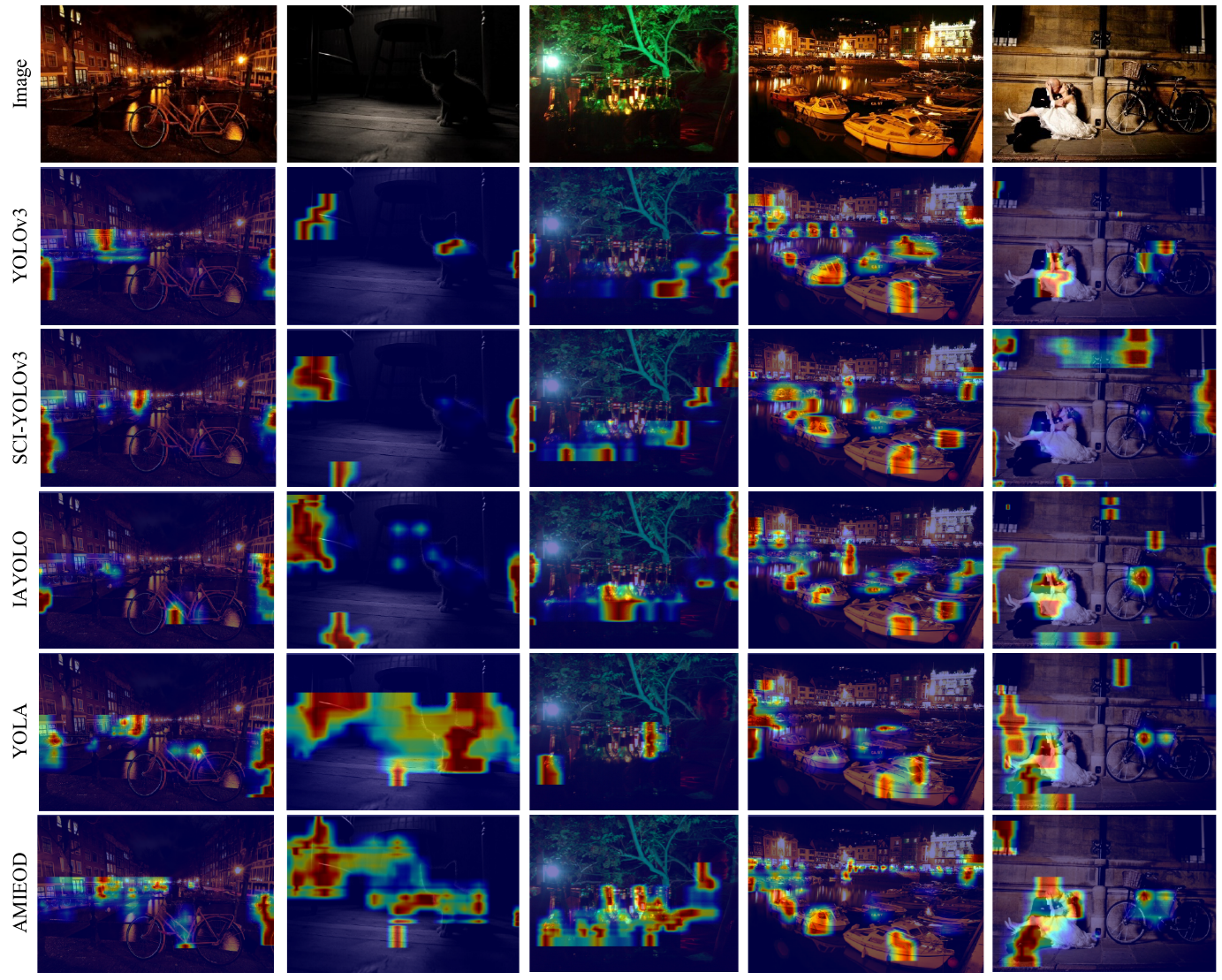}
		\caption{Comparison of heatmap visualization results. All comparison algorithms adopt YOLOv3 as base detector.}
		\label{heat_vis}
	\end{figure*}

	To interpret the attention mechanisms of different detection models, we present heatmap visualization results of decoder features using HiResCAM \cite{draelos2020use} in Fig. \ref{heat_vis}. 
	Compared to the baseline YOLOv3, SCI-YOLOv3 fails to improve attention to target regions. Instead, it often exhibits increased focus on irrelevant background areas. This may be due to artifacts introduced during the image enhancement process, which can mislead the detection model into mistakenly interpreting non-target regions as containing objects.
	The results of IAYOLO and YOLA demonstrate improved attention toward target areas compared to the baseline. However, they still struggle to capture many subtle or less distinguishable targets, particularly in complex or low-contrast scenes.
	In contrast, the proposed AMIEOD exhibits a more concentrated and accurate focus on actual target regions within low-illumination images. Notably, it demonstrates a superior ability to attend to small, dimly lit, and partially occluded objects that are typically challenging for conventional detectors. These results highlight AMIEOD’s effectiveness in learning illumination-invariant and semantically meaningful representations, thereby enhancing detection performance in adverse lighting conditions.

\subsubsection{Validity on Different Baselines}

\begin{figure}[t]
    		\centering
    		\includegraphics[scale=0.66]{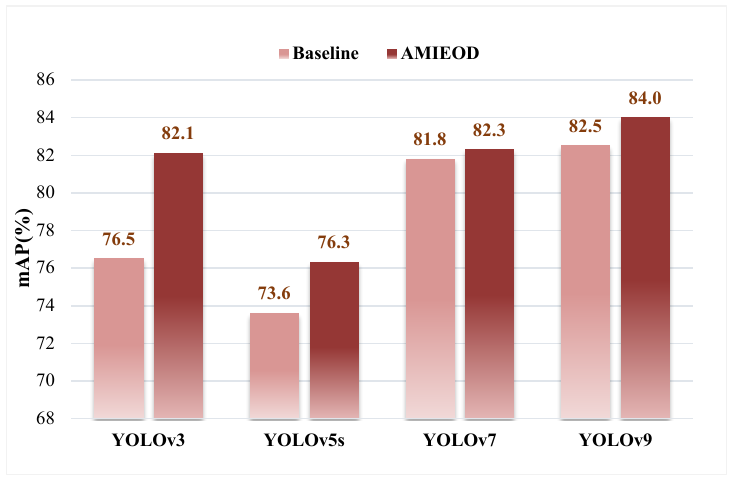}
    		\caption{The performances on different baselines.}
    		\label{various_baseline}
	\end{figure}

	Our proposed AMIEOD is a general image preprocessing method designed to enhance detection performance in low-illumination images, and it is not confined to any specific detection model. As illustrated in Fig. \ref{various_baseline}, the proposed method consistently improves detection performance on the ExDark dataset across various versions of YOLO detector, including YOLOv3, YOLOv5s, YOLOv7, and the Gelan-s model of YOLOv9. Specifically, AMIEOD achieves mAP improvements of 5.6$\%$, 2.7$\%$, 0.5$\%$, and 1.5$\%$, respectively.
	These results demonstrate the generalizability and effectiveness of the proposed approach on different detection algorithms under low-illumination conditions.

\subsubsection{Generalization on Various Datasets}
	To demonstrate the generalization capability of the proposed method, this section conducts comparative experiments using visible images from DarkFace, LLVIP, and M3FD datasets. These datasets differ significantly in terms of object categories, scene complexity, and lighting conditions. 
	\par 
    The detailed results are presented in Table \ref{dataset_generalization}. Since Darkface and LLVIP are single-category datasets, their AP are equivalent to mAP. It can be observed that our proposed AMIEOD consistently improves detection performance over the baseline and outperforms existing methods across different datasets. Specifically, on Darkface and LLVIP, AMIEOD achieves the highest AP metric, surpassing the second-best method by 0.1\% and 0.4\%, respectively. For the multi-category M3FD dataset, AMIEOD obtains the highest AP in four out of six categories, and achieves a 0.7\% mAP improvement over the baseline YOLOv3. Moreover, compared with other advanced variants such as SCI-YOLOv3, IAYOLO, and YOLA, AMIEOD still demonstrates superior overall performance, confirming its better generalization capability and robustness in dark scenarios.
    \par 
    In terms of precision and recall, AMIEOD demonstrates a well-balanced performance. On Darkface, it achieves the best AP with 80.0\% precision and 58.1\% recall, indicating a favorable trade-off compared with other methods. On LLVIP, AMIEOD attains the highest recall 87.6\%, while maintaining competitive precision of 87.6\%. On M3FD, it further achieves a competitive precision of 91.1\% and the highest recall of 85.4\% among all methods. These results confirm that AMIEOD effectively improves recall, thereby enhancing the mAP metric, which indicates our method is more comprehensive and reliable in challenging low-illumination scenarios.
    \par 
    It is worth noting that, on the M3FD dataset, most categories (e.g., car, bus, motorbike, lamp, and truck) achieve high AP (around 90\%) across different methods,, while people remains more challenging (around 80\%). This is mainly because M3FD is a visible–infrared dataset with shared annotations, where many pedestrian instances are prominent in infrared images but weakly observable in visible images. Since only visible images are used in our experiments, this modality mismatch leads to relatively lower AP for the people category.
	\begin{table*}[ht]
		\centering 
		\caption{Detailed results on the Darkface, LLVIP, and M3FD datasets. The best result is highlighted in bold.  ``P'' refers to precision rate. ``R'' indicates recall rate. We report the average precision (AP) for all categories in M3FD, from left to right: People, Car, Bus, Motorbike, Lamp, Truck. }
		\label{dataset_generalization} 	
		\setlength{\tabcolsep}{2mm}
			\begin{threeparttable}
				\begin{tabular}{c|c|ccc|ccc|ccccccccc}  
					\toprule
					\multirow{2}*{Method} &\multirow{2}*{Type} &&Darkface&   &&LLVIP &&&&&&M3FD&&&&\\
					\cline{3-17}
					& &P&R&AP	&P&R&AP &Peo.&Car&Bus&Mot.&Lam.&Tru.&P&R&mAP\\
					\midrule			
					YOLOv3 &   -    &78.4&57.9&63.0&
					93.3&84.8&91.4&
					80.6&94.2&89.5&\textbf{95.3}&86.7&89.3&91.2&84.9&89.3\\ 				
					SCI-YOLOv3 &LLIE&78.7&58.4&63.4&
					92.6&84.5&91.3&
					79.6&93.8&90.2&94.3&84.0&85.0&89.0&83.5&87.8\\
					SCI-YOLOv3*& JED&79.0&\textbf{58.8}&64.2
					&93.0&85.5&92.4
					&80.6&94.2&90.4&94.6&86.1&86.8&\textbf{91.3}&83.7&88.8				\\
					IAYOLO     & JED&78.4&58.5&63.9
					&93.1&85.2&91.6
					&\textbf{80.9}&94.3&90.1&94.4&85.4&86.7&89.8&85.1&88.6				\\  
					YOLA       &IIFE&\textbf{80.3}&58.4&63.9
					&\textbf{94.0}&86.6&92.1
					&80.4&94.2&90.6&94.8&87.4&87.7&91.0&85.4&89.2	\\
					\midrule
					\rowcolor{gray!20}\textbf{AMIEOD(Our)}&-&80.0&58.1&\textbf{64.3} 
					&89.3&\textbf{87.6}&\textbf{92.8}
					&80.7&\textbf{94.6}&\textbf{90.9}&94.6&\textbf{90.0}&\textbf{89.3}&91.1&\textbf{85.4}&\textbf{90.0} \\
					\bottomrule
				\end{tabular}
				
			\end{threeparttable}
	\end{table*}

	
	\subsubsection{Comparison with Ground Truth}

\begin{figure}[ht]
	\centering
	\includegraphics[scale=0.35]{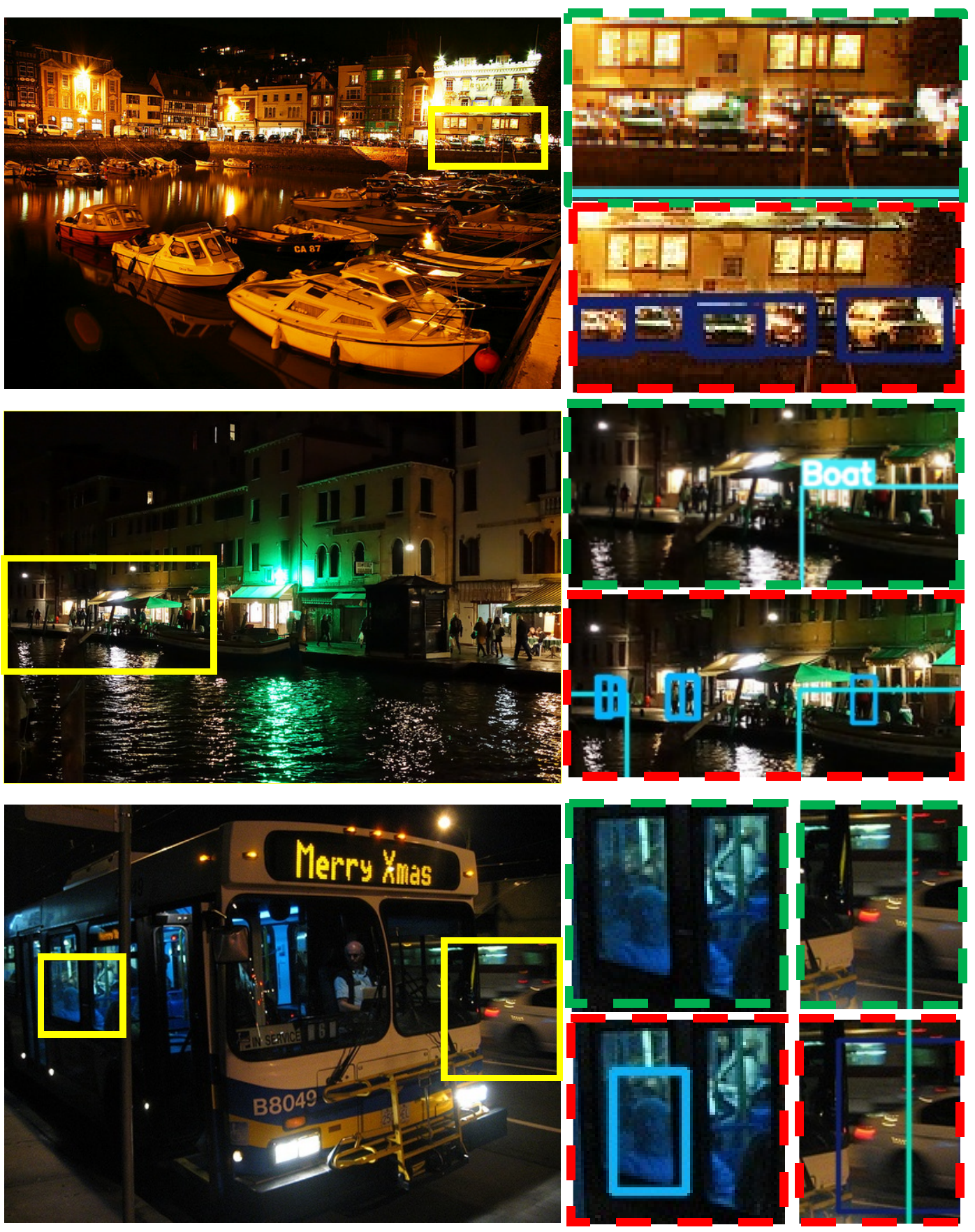}
	\caption{Comparison between our detection results and the ground truth. The yellow boxes denote enlarged regions from the original images. The green dashed boxes indicate the visual ground truth within the enlarged regions, including unlabeled targets. The red dashed boxes represent the detection results of our AMIEOD within the enlarged regions.}
	\label{ground truth compare}
	\end{figure}

	Fig. \ref{ground truth compare} presents the detection results of the proposed AMIEOD method in comparison with the ground truth. Some targets are correctly detected but are not annotated in ground truth. Such detections are counted as false positives, which leads to a decrease in Precision (P) metric. However, these targets correspond to valid objects and would be categorized as true positives if the ground-truth annotations were complete, in which case the precision score would be correspondingly higher.
	\par 
	Consequently, as shown in Tables \ref{Compared table} and \ref{dataset_generalization}, these detections are mistakenly treated as false positives under the current evaluation protocol, resulting in lower Precision (P) and a negative impact on the mAP metric. It should be noted that this limitation stems from annotation incompleteness rather than detection algorithm. This highlights the inherent limitations of existing evaluation metrics in adverse scenarios, where exhaustive annotation is particularly challenging. Meanwhile, the ability of AMIEOD to detect such unlabeled yet valid targets also indicates its strong generalization capability and reduced tendency to overfit.

	
	
	\subsection{Ablation Study}
	\subsubsection{Ablation Experiment Setting}
	We conduct comprehensive ablation studies on Exdark to evaluate the impact of the proposed MEIEM, DGRL, ESM, and DGCE. 
	First,to fully verify the validity of each sub-module in the MEIEM and ESM optimized by DGCE,  we adopt YOLOv3 as the baseline and progressively integrate the components of MEIEM including PIEM, JIEM, and IAEM.
	The detailed model configurations are summarized in Table \ref{ablation experiments setting}. Specifically, V1$\sim$V8 progressively incorporate different sub-modules of the proposed MEIEM to evaluate their individual and combined effects. V9 and V10 further build upon the complete MEIEM. V9 introduces the ESM optimized with the DGCE loss, while V10 incorporates the DGRL for detection-aware enhancement learning. Finally, V11 represents the complete AMIEOD framework, integrating MEIEM, DGRL, and ESM with DGCE.
	\begin{table}[!t]
		\caption{Ablation experiments setting.}
		
		\label{ablation experiments setting} 	
		\centering 
		\setlength{\tabcolsep}{1.9mm}
			\begin{threeparttable}
				\begin{tabular}{c|c|ccc|c|c}  
					\toprule
					&\multirow{2}*{YOLOv3}&&MEIEM&&\multirow{2}*{DGRL}&ESM\\
					\cline{3-5}
					&&PIEM&JIEM&IAEM&&(DGCE)\\
					\midrule
					V1&\Checkmark&&&&&\\
					V2&\Checkmark&\Checkmark&&&&\\
					V3&\Checkmark&&\Checkmark&&& \\
					V4&\Checkmark&&&\Checkmark&&\\
					V5&\Checkmark&\Checkmark&\Checkmark&&&  \\
					V6&\Checkmark&\Checkmark&&\Checkmark&&  \\
					V7&\Checkmark&&\Checkmark&\Checkmark&&  \\
					V8&\Checkmark&\Checkmark&\Checkmark&\Checkmark&&   \\
					V9&\Checkmark&\Checkmark&\Checkmark&\Checkmark&&\Checkmark   \\
					V10&\Checkmark&\Checkmark&\Checkmark&\Checkmark&\Checkmark&   \\
					V11&\Checkmark&\Checkmark&\Checkmark&\Checkmark&\Checkmark&\Checkmark \\
					\bottomrule
					
				\end{tabular}
			\end{threeparttable}
		
	\end{table}

	\par 
	
	\subsubsection{Ablation Experiment Results}

	\begin{figure}[ht]
		\centering
		\includegraphics[scale=0.66]{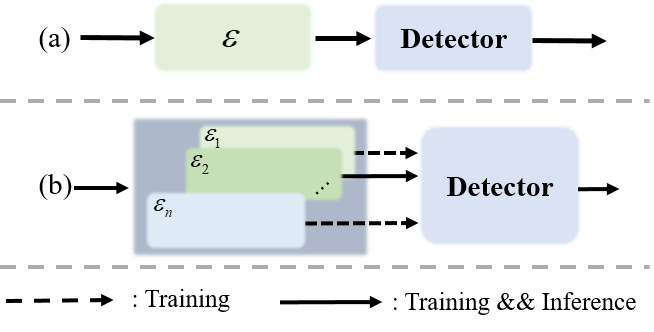}
		\caption{Single- and multi-enhancement training strategies. (a) Single-enhancement method, where training is conducted using images enhanced by a single enhancement model. (b) Multi-enhancement method, where training leverages both the original image and multiple enhanced variants.}
		\label{one_multi_input}
	\end{figure}	
	
	\par
	
	\begin{table}[!t]
		\caption{Ablation experiments results on Exdark. ``Set.'' is the model setting. ``Inf.'' indicates preprocessing method when inferring. ``Ori.'' indicates original input is detected without image enhancement.}
		
		\label{ablation experiments} 	
		\centering 
		\setlength{\tabcolsep}{1.1mm}
			\begin{threeparttable}
				\begin{tabular}{c|c|ccc|c|c|c|c}  
					\toprule
					\multirow{2}*{\diagbox{Set.}{Inf.}}&\multirow{2}*{Ori.}&&$\mathcal{E}$&&mAP&Params&GFLOPs&\multirow{2}*{FPS}\\
					\cline{3-5}
					&&PIEM&JIEM&IAEM&($\%$) & (M)  &(G)&\\
					\midrule
					V1&\Checkmark&&&&76.5  & 61.5567  & 77.36 &103\\
					\midrule
					V1+PIEM&&\Checkmark&&&76.6& 61.5570  & 77.58 & 101\\
					V1+JIEM&&&\Checkmark&&77.4& 61.5570  & 77.65 & 100\\
					V1+IAEM&&&&\Checkmark&77.8& 61.7216  & 77.40 & 56\\
					
					\midrule
					\multirow{2}*{V2}&\Checkmark&&&&76.9&\multirow{2}*{61.5570} & 77.36 & 103 \\
					&&\Checkmark&&&76.2&						& 77.58 & 100  \\
					\midrule       
					\multirow{2}*{V3}&\Checkmark&&&&78.1&\multirow{2}*{61.5570} & 77.36 & 104 \\
					&&&\Checkmark&&78.1&						& 77.65 & 101 \\
					\midrule   
					
					\multirow{2}*{V4}&\Checkmark&&&&77.9&\multirow{2}*{61.7216} & 77.36 & 103 \\
					&&&&\Checkmark&78.0&						& 77.40 &  55  \\
					\midrule   
					
					\multirow{3}*{V5}&\Checkmark&&&&77.4&\multirow{3}*{61.5573} & 77.36 &104 \\
					&&\Checkmark&&&75.7&  						& 77.58 &99\\
					&&&\Checkmark&&77.6&  						& 77.65 &100\\
					\midrule	
					
					\multirow{3}*{V6}&\Checkmark&&&&77.1&\multirow{3}*{61.7219} & 77.36 &103\\
					&&\Checkmark&&&77.2&						& 77.58 &101\\
					&&&&\Checkmark&77.6&						& 77.40 & 56\\
					\midrule
					\multirow{3}*{V7}&\Checkmark&&&&78.3&\multirow{3}*{61.7219} & 77.36 & 104\\
					&&&\Checkmark&&78.6&						& 77.65 & 100\\
					&&&&\Checkmark&78.3&						& 77.40 & 57\\
					\midrule
					
					\multirow{4}*{V8}&\Checkmark&&&&78.3&\multirow{4}*{61.7222} & 77.36 &103\\
					&&\Checkmark&&&77.4&  						& 77.58 &99\\
					&&&\Checkmark&&78.9&  						& 77.65 &99\\
					&&&&\Checkmark&78.8&   					& 77.40 &57\\
					\midrule				 
					V9				 &-&&-&		   &79.0&85.2384				&  -	&  75 \\
					\midrule
					\multirow{4}*{V10}&\Checkmark&&&&81.9&\multirow{4}*{61.7222} & 77.36 &103\\
					&&\Checkmark&&&79.9&  						& 77.58 &99\\
					&&&\Checkmark&&82.0&  						& 77.65&99\\
					&&&&\Checkmark&82.0&  						& 77.40&56\\
					\midrule				 
					V11				 &-&&-&&\textbf{82.1}&85.2384& - &70 \\
					\bottomrule
					
				\end{tabular}
			\end{threeparttable}
		
	\end{table}	
	
	Table \ref{ablation experiments} reports the ablation results on the ExDark dataset to evaluate the effectiveness of different model settings listed in Table \ref{ablation experiments setting}.
	V1 represents the baseline detector using YOLOv3, achieving an mAP of 76.5\%. As shown in Fig. \ref{one_multi_input} (a), the configurations denoted as V1 + (PIEM / JIEM / IAEM) introduce a single enhancement strategy (PIEM, JIEM, or IAEM) as an independent preprocessing module. These settings bring moderate performance gains over the baseline, achieving mAP of 76.6\%, 77.4\%, and 77.8\% respectively, indicating that single-enhancement network can partially mitigate low-illumination degradation. However, the improvement remains limited, as the single-enhancement models are struggled to adapt to complex low-illumination environments.
	As illustrated in Fig. \ref{one_multi_input}(b), V2$\sim$V8 and V9 incorporate multiple enhancement models during the training stage, while only a single enhancement strategy is selected during inference.
	\par 
	V2$\sim$V4 correspond to the model settings in Table IV. Unlike V1 + (PIEM / JIEM / IAEM), both the original image and the enhanced image are used to compute detection loss for optimization. As a result, compared with single-enhancement method, these jointly trained with original image achieve more stable and consistent improvements, demonstrating that multi-enhancement is more effective than independent preprocessing.
	\par
	V5$\sim$V7 further integrate multiple enhancement experts into MEIEM, enabling the detector to learn from diverse enhancement perspectives during training. Compared with V2 $\sim$ V4, these variants achieve higher mAP, indicating that different enhancement strategies provide complementary cues. V8 incorporates PIEM, JIEM, and IAEM into a unified MEIEM, achieving relatively higher mAP. When JIEM is selected as the enhancement strategy during inference, V8 achieves an mAP of 78.9\%, outperforming all preceding configurations. This improvement demonstrates that integrating suitable and diverse preprocessing modules within MEIEM allows the detector to exploit image information more comprehensively.
	\par 
	V9 and V10 introduce detection-guided optimization mechanisms on top of the complete MEIEM. V9 incorporates the ESM trained with the proposed DGCE loss, enabling adaptive selection of a suitable enhancement strategy for each input image. This results in a further performance improvement, achieving an mAP of 79.0\%.
	Building upon V8, V10 further introduces the DGRL, which directly supervises enhancement learning using detection loss. By more tightly aligning enhancement optimization with the detection task, V10 achieves the best overall performance, with significant mAP improvements observed across all four inference settings. Specifically, compared with V8, the introduction of DGRL leads to mAP increases of 3.6\%, 2.5\%, 3.1\%, and 3.2\% under the original input and the three enhanced conditions of PIEM, JIEM, and IAEM, respectively.
	
	\par 
	V11, built upon V10, further introduces the ESM optimized with the proposed DGCE loss, enabling the model to adaptively select the most suitable enhancement preprocessing strategy during inference. As a result, V11 achieves a state-of-the-art mAP of 82.1\%.
	\par 
	Overall, the ablation results verify that jointly optimized multi-expert enhancement and detection-guided learning are both essential for achieving robust performance in low-illumination object detection.
	
	

	\par 

	\subsubsection{Hyperparameter Ablation}
	
	\begin{figure*}[ht]
		\centering
		\includegraphics[scale=0.56]{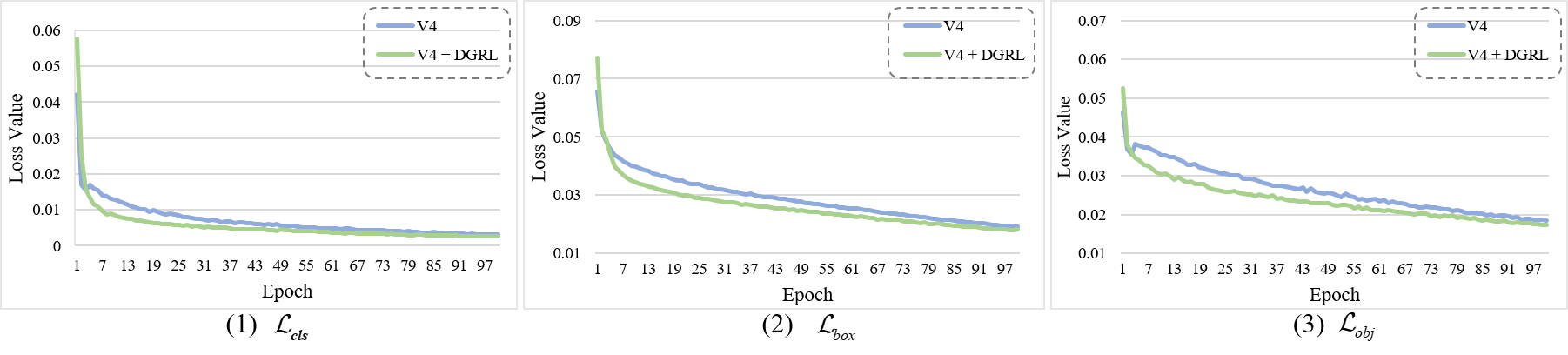}
		\caption{Detection loss trends during MEIEM and detector training. Horizontal axis refers the number of epoch. Vertical axis is the loss value of $\mathcal{L}_{cls}$, $\mathcal{L}_{box}$, $\mathcal{L}_{obj}$. V4 indicates the model setting including the MEIEM and base detector YOLOv3. }
		\label{det_loss3}
	\end{figure*}	
	
	To evaluate the impact of coefficient variations in $\mathcal{L}_{\text{stage1}}$, we compare results for different values of the weighting hyperparameter $\alpha$ in Table \ref{ablation of coefficient}. The best performance occurs at $\alpha = 0.2$, with accuracy declining as $\alpha$ increases when $\alpha$ exceeds 0.2.
	
	\begin{table}[h]
	\caption{Ablation experiments of different coefficient $\alpha$.}
	\label{ablation of coefficient} 	
	\centering 
	\small
	\renewcommand{\arraystretch}{1}
	\setlength{\tabcolsep}{2.2mm}
	\begin{threeparttable}
		\begin{tabular}{c|ccccccc}  
		\toprule
		$\alpha$ &0.1  &\textbf{\textcolor{orange}{0.2}}&0.3&0.4&0.5&0.6&0.7\\
		\midrule
		
		mAP($\%$) & 80.7 &\textbf{\textcolor{orange}{82.1}}&80.8&79.9&79.1&78.5&78.2\\
		\bottomrule
		\end{tabular}
		
		\end{threeparttable}	

	\end{table}

	\subsubsection{Detailed Analysis of DGRL}
	
	
	To analyze the effectiveness and stability of the proposed DGRL, we visualize the evolution of different detection loss components during training, including the classification loss $\mathcal{L}_{cls}$, bounding box regression loss $\mathcal{L}_{box}$, and objectness loss $\mathcal{L}_{obj}$. As shown in Fig.~\ref{det_loss3}, we compare the loss curves of model setting V4 and V4+ DGRL. Overall, introducing DGRL leads to faster convergence and consistently lower loss across all components.
	\par 
	In the early training stage, DGRL notably accelerates loss reduction, particularly for $\mathcal{L}_{cls}$ and $\mathcal{L}_{obj}$, indicating that using the enhancement output with minimum detection loss as the regression target provides a more detection-aligned supervisory signal. In later stages, the loss curves with DGRL exhibit smoother and more stable convergence.
	Moreover, the consistent decrease in $\mathcal{L}_{box}$ shows that DGRL also improves localization accuracy rather than favoring only easier classification cases. Since the regression target is dynamically selected and detached from gradient propagation, DGRL avoids reinforcing noisy gradients and acts as a relative risk minimization mechanism among multiple enhancement candidates, thereby stabilizing training.
	
	


	\subsection{Detailed Analysis of ESM and DGCE}
\begin{figure}[ht]
		\centering
		\includegraphics[scale=0.55]{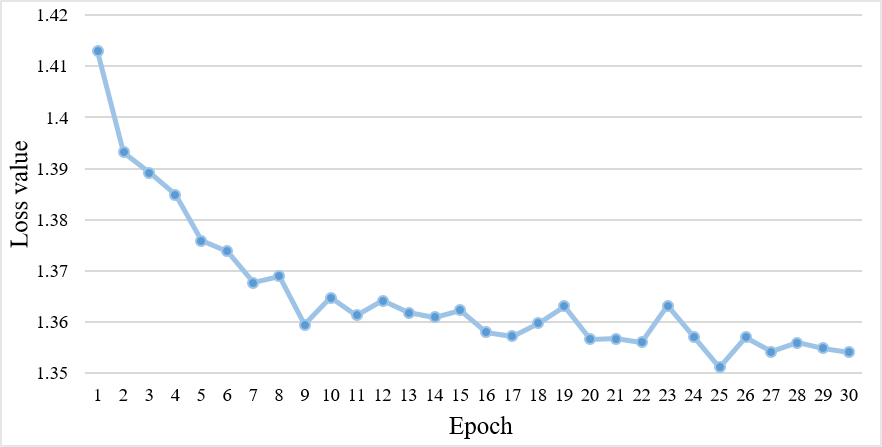}
		\caption{Variation of the DGCE loss during ESM training stage with 30 epochs. Horizontal axis refers the number of epoch. Vertical axis is the loss value. } 
		\label{DGCE curve}
	\end{figure}

	Fig. \ref{DGCE curve} illustrates the training loss curve of the ESM optimized with the proposed DGCE loss. The loss decreases rapidly in the early stage and gradually converges to a stable value, indicating efficient and stable optimization. Although slight fluctuations are observed in later iterations, the overall downward trend suggests that the detection-guided pseudo-labels provide a reliable supervisory signal.
	
	As reported in Table \ref{compare_ESM}, using a single fixed preprocessing strategy during inference leads to suboptimal performance. Specifically, directly feeding the original images achieves an mAP of 81.9\%, while applying PIEM, JIEM, and IAEM individually results in mAPs of 79.9\%, 82.0\%, and 82.0\%, respectively. These results are reported as overall performance on the test set; however, for individual input samples, the most suitable enhancement strategy varies, and no single preprocessing method is optimal for all images.	
	\par
	
	We randomly conduct expert selection, with mAP values ranging from 81.3\% to 81.6\% across three independent runs (R1$\sim$R3), which are consistently lower than those of the best fixed expert. In contrast, the proposed ESM-based adaptive routing achieves the highest mAP of 82.1\%, outperforming both fixed expert selection and random routing. These results demonstrate that the learned expert selection policy is more effective than static or stochastic strategies, even when individual experts already provide strong baselines.
	
	
	
	\begin{table}[h]
		\caption{
		Comparison of different inference settings. ``Inf.'' denotes the preprocessing method used during inference. ``R'' indicates random selection from the original image, PIEM, JIEM, and IAEM.}
		\label{compare_ESM} 	
		\centering 
		\small
		\renewcommand{\arraystretch}{1}
		\setlength{\tabcolsep}{1.5mm}
			\begin{threeparttable}
				\begin{tabular}{c|cccccccc}  
					\toprule
					Inf. & Ori.  &PIEM	&JIEM	&IAEM	&R1	&R2	&R3	& Our\\
					\midrule
					
					mAP($\%$) & 81.9 &79.9	&82.0	&82.0	&	81.3	&81.4	&81.6	&	\textbf{82.1}\\
					\bottomrule
				\end{tabular}
				
			\end{threeparttable}	
	\end{table}
	
	\subsection{Model Complexity Analysis}

		Table~\ref{ablation experiments} provides a detailed analysis of model complexity for each sub-module, including parameter count, computational cost (GFLOPs), and inference speed (FPS), with experimental settings summarized in Table~\ref{ablation experiments setting}. Specifically, different preprocessing strategies in MEIEM introduce varying but negligible computational overhead. Incorporating individual enhancement modules keeps the parameter count around 61.56$\sim$61.72M and increases GFLOPs only slightly from 77.36G to 77.65G, while maintaining real-time inference speed. Moreover, when optimized with the proposed DGRL (V10 vs. V8), the model achieves notable performance gains without additional computational cost. The complete AMIEOD framework (V9 \& V11) further introduces the ESM, increasing parameters to 85.24M, yet still achieves 70 FPS, demonstrating a favorable trade-off between detection accuracy and efficiency for real-time applications.
		\par 
		Table VIII compares the model complexity of several representative enhancement-based detection methods. SCI-YOLOv3 introduces negligible parameter and FLOPs overhead (+0.0003M / +0.22G) and thus maintains an inference speed comparable to the baseline YOLOv3 (101 FPS vs. 103 FPS). IAYOLO and YOLA incur relatively small parameter and FLOPs increases (+0.1649M / +0.04G and +0.0086M / +3.62G, respectively), but their inference speed is noticeably reduced due to additional image filtering, resulting in 56 FPS for IAYOLO and 67 FPS for YOLA.
		\par 
		
		In contrast, although AMIEOD introduces additional parameters and computation due to the ESM, this overhead is mainly introduced by the ESM itself. Specifically, as shown in Table~\ref{ablation experiments}, the ESM accounts for 23.5162M of the additional 23.6817M parameters. 
		Moreover, as reported in Table~\ref{Complexity_compare}, when detecting images using the original input, the entire additional 5.40G computational cost is attributed to the ESM. 
		This overhead is necessary to enable expert selection under complex illumination conditions. Even so, AMIEOD delivers substantial accuracy gains while maintaining 70 FPS at an input size of 640$\times$640, meeting real-time requirements in most scenarios and demonstrating a favorable performance–efficiency trade-off for low-illumination detection.
	\begin{table}[h]
		\caption{Comparison of model complexity.}
		\centering 
		
		\renewcommand{\arraystretch}{1}
		\setlength{\tabcolsep}{3mm}{
			\begin{threeparttable}
				\begin{tabular}{c|ccc}  
					\toprule
					Method	&Params(M)$\downarrow$&GFLOPs(G)$\downarrow$&FPS$\uparrow$ \\
					\midrule
					YOLOv3		& 61.5567	&	77.36	&	103	\\
					SCI-YOLOv3	& +0.0003	&	+0.22	&	101	\\
					IAYOLO		& +0.1649	&	+0.04	&	56	\\
					YOLA		& +0.0086	&	+3.62	&	67	\\
					AMIEOD(Our) & +23.6817	&	+(5.40$\sim$5.62)		&	70	\\
					
					%
					
					\bottomrule
					
				\end{tabular}
		\end{threeparttable}	}
		\label{Complexity_compare} 	
	\end{table}

	\subsection{Analysis of False Cases}
	
\begin{figure}[ht]
		\centering
		\includegraphics[scale=0.56]{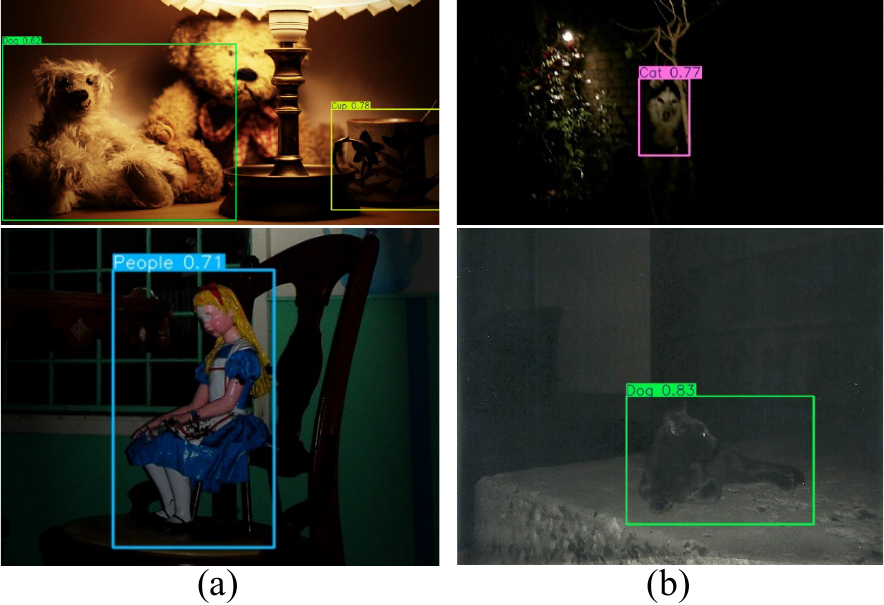}
		\caption{Visualization results of false case.} 
		\label{false case}
	\end{figure}

	Despite the overall performance improvements, AMIEOD still encounters challenges in discriminating camouflaged or visually ambiguous objects. As shown in Fig.~\ref{false case}(a), objects with anthropomorphic or animal-like appearances (e.g., dolls) may be incorrectly detected as real dogs or persons under extremely low-illumination conditions, where the detector tends to rely on coarse structural cues.
	Fig.~\ref{false case}(b) further illustrates occasional category confusion between visually similar classes, such as dogs and cats, caused by weakened discriminative details under low-illumination degradation. 
	These failure cases highlight the need for further improvements in fine-grained semantic modeling and stronger supervision, such as prototype-based contrastive learning with hard-negative mining, to better separate highly similar categories.
	
	\section{Conclusion}
	
	In this study, we propose a detector-flexible object detection approach for low-illumination image, denoted by AMIEOD, 
	It first introduces MEIEM to enhance the image from four distinct perspectives, thereby comprehensively exploiting image information for model training to improve detection performance in poorly lit images. 
	To better align MEIEM with the detector, we design DGRL, which computes a regression loss to optimize MEIEM in a detection-guided manner.
	Additionally, we propose ESM with DGCE loss to adaptively select the optimal enhancement strategy from MEIEM for each input. 
	Extensive experiments demonstrate superiority over existing approaches in object detection for low-illumination scenarios.	
	\par 
	
	
	
	While the expert selection mechanism introduces additional complexity, the inference efficiency remains acceptable for real-time applications. Future work will explore better efficiency–performance trade-offs. In addition, incorporating fine-grained semantic modeling could further improve category discrimination under low-illumination conditions, especially for visually ambiguous or camouflaged objects.

\bibliographystyle{IEEEtran}
\bibliography{mybibfile_new}

\vfill
\end{document}